\documentclass[runningheads]{llncs}
\usepackage[T1]{fontenc}
\usepackage{graphicx}
\usepackage{amsmath}
\usepackage{amssymb}
\usepackage{mathtools}
\usepackage[textsize=tiny]{todonotes}
\usepackage{cleveref}
\renewcommand{\P}{\mathcal{P}}
\newcommand{\R}{\mathbb{R}}

\newcommand{\N}{\mathcal{N}}
\renewcommand{\d}{\mathrm{d}}

\DeclareMathOperator{\tr}{tr}
\DeclareMathOperator{\Exp}{Exp}
\DeclareMathOperator{\Log}{Log}
\DeclareMathOperator{\WG}{WG}
\DeclareMathOperator*{\argmin}{arg\,min}

\begin{document}
\title{A probabilistic view on Riemannian machine learning models for SPD matrices}

\titlerunning{Probabilistic view on tools used for SPD matrices}
%
\author{Thibault de Surrel\inst{1}\orcidID{0009-0001-5341-1538} \and
Florian Yger\inst{2}\orcidID{0000-0002-7182-8062} \and Fabien Lotte\inst{3}\orcidID{0000-0002-6888-9198} \and Sylvain Chevallier\inst{4}\orcidID{0000-0003-3027-8241}}
\authorrunning{T. de Surrel et al.}
%
\institute{LAMSADE, CNRS, PSL Univ. Paris-Dauphine, France \email{thibault.de-surrel@lamsade.dauphine.fr}\and
LITIS, INSA Rouen-Normandy,  Rouen, France \and 
Inria center, University of Bordeaux / LaBRI, Talence, France\and
TAU, LISN, University Paris-Saclay, France}

\maketitle 

%
\begin{abstract}
The goal of this paper is to show how different machine learning tools on the Riemannian manifold $\P_d$ of Symmetric Positive Definite (SPD) matrices can be united under a probabilistic framework. For this, we will need several Gaussian distributions defined on $\P_d$. We will show how popular classifiers on $\P_d$ can be reinterpreted as Bayes Classifiers using these Gaussian distributions. These distributions will also be used for outlier detection and dimension reduction. By showing that those distributions are pervasive in the tools used on $\P_d$, we allow for other machine learning tools to be extended to $\P_d$.

\end{abstract}
\section{Introduction}

Symmetric Positive Definite (SPD) matrices appear in several applications of Machine Learning (ML), such as in Brain Computer Interfaces \cite{ygerRiemannianApproachesBrainComputer2017}, biomedical image analysis \cite{pennecManifoldvaluedImageProcessing2020} or video processing \cite{Tuzel2008}. More precisely, the Riemannian structure of the set of $d \times d$ SPD matrices, denoted $\P_d$, has been leveraged to develop a range of tools to deal with data lying on this manifold. In this work, we use the affine invariant Riemannian framework \cite{pennecManifoldvaluedImageProcessing2020}, built from the following Riemannian metric on $\P_d$, called the Affine Invariant Riemannian Metric (AIRM) and defined on the tangent space $T_P \P_d$ at a point $P \in \P_d$ by:
\begin{equation}
    \label{def:AIRM_metric}
    \forall U,V \in T_P \P_d,~ \langle U, V \rangle_P = \tr(P^{-1}UP^{-1}V).
\end{equation}
Using the geodesic induced by this geometry, one can define the AIRM distance between two points $P,Q \in \P_d$ as:
$$\delta(P,Q) = \|\log(P^{-1/2}QP^{-1/2})\|_F$$
The goal of this paper is to revisit some ML tools used on $\P_d$ with a probabilistic lens. More precisely, we will show how different probability distributions on $\P_d$ can be used to reinterpret classifiers, tools detecting outliers or performing dimension reduction. We will show that all these tools can be united under the same probabilistic framework. For this, we will start by describing, in \cref{sec:prob_distrib}, the different probability distributions on $\P_d$ that we will use. In \cref{sec:classification}, popular classifiers on $\P_d$ will be reinterpreted using those distributions, as well as an outlier detection tool in \cref{sec:outlier_detection} and dimension reduction tools in \cref{sec:dim_reduction}. 

\section{Different probability distributions on the manifold of SPD matrices}
\label{sec:prob_distrib}

\subsection{The isotropic Gaussian distribution}
\label{sec:isotropic_gaussian}
The authors of \cite{saidRiemannianGaussianDistributions2016} define an isotropic Gaussian distribution $G(\bar{X}, \sigma^2)$ on $\P_d$ by its probability density function, with respect to the Riemannian volume:
$$p_{\bar{X}, \sigma}(X) = \frac{1}{\zeta(\sigma)} \exp\left(-\frac{\delta(X, \bar{X})^2}{2\sigma^2}\right).$$
This distribution depends on two parameters: $\bar{X} \in \P_d$ acting as the center of mass of the distribution and $\sigma > 0$ acting as the spread of the distribution. The authors also give an exact expression of the normalization constant $\zeta(\sigma)$ that does not depend on $\bar{X}$, but only on $\sigma$. Using proposition 7 of \cite{saidRiemannianGaussianDistributions2016}, one can estimate the two parameters $\bar{X}$ and $\sigma$ of the distribution from an independent sample $X_1,...,X_N \sim G(\bar{X}, \sigma)$ using the Maximum Likelihood Estimator (MLE): for $\bar{X}$, the MLE $\hat{X}_N$ is the Riemannian mean of the sample (see \cite{moakherDifferentialGeometricApproach2005}) and for $\sigma$, the MLE is the only solution of the following non-linear equation in $\sigma$:
$$\sigma^3 \frac{\d}{\d \sigma} \log \zeta(\sigma) = \frac{1}{N}\sum_{i=1}^N \delta(X_i, \hat{X}_N)^2.$$ 

\subsection{Wrapped distributions}
\label{sec:wrapped_gaussian}
We now share a way to define an anisotropic Gaussian distribution on the manifold $\P_d$ using the concept of \emph{wrapped Gaussian} (WG) distribution \cite{chevallierExponentialWrappedDistributionsSymmetric2022}. This distribution depends on three parameters: $P \in \P_d$, $\mu \in \R^{d(d+1)/2}$ and $\Sigma \in \P_{d(d+1)/2}$. Then, we say that the random variable $\mathbf{X}$ on $\P_d$ follows a wrapped Gaussian distribution $\WG(P; \mu, \Sigma)$ if:

$$\mathbf{X} = \Exp_{P}(\mathbf{t}),~\mathbf{t}\sim\N(\mu, \Sigma)$$ 

The wrapped Gaussian corresponds to a multivariate Gaussian distribution $\N(\mu, \Sigma)$ in normal coordinates at $P$. This distribution is further analyzed in \cite{desurrel2025wrappedgaussianmanifoldsymmetric}. In particular, the density of $\WG(P; \mu, \Sigma)$ is given by:
$$f_{P;\mu,\Sigma}(X) = \frac{g_{\mu,\Sigma}(\Log_{P}(X))}{|J_{P}(\Log_{P}(X))|} $$
where $g_{\mu, \Sigma}$ is the density of the multivariate Gaussian $\mathcal{N}(\mu, \Sigma)$ and $J_{P}(\cdot ) = \det(\mathrm{d} \Exp_{P}(\cdot))$ is the Jacobian determinant of the exponential map $\Exp_{P}$. The closed-form formula of the Jacobian determinant $J_{P}$ is given in Proposition 4.3 of \cite{desurrel2025wrappedgaussianmanifoldsymmetric}. Moreover, the authors show that the parameters $(P, \mu, \Sigma)$ of the WG can be estimated from a sample $X_1,...,X_N$ using the method of moments in the case of $\mu = 0$ \emph{a priori} and using an MLE in a general setting where $\mu \neq 0$.


\subsection{Other probability distributions on $\P_d$}
An alternative definition of an anisotropic Gaussian distribution on $\P_d$ is given in \cite{pennecIntrinsicStatisticsRiemannian2006}. This approach maximizes entropy given a mean and covariance matrix (see Theorem 13.2.2 of \cite{kagan1973characterization}). On $\P_d$, the entropy-maximizing distribution with mean $\bar{X}$ and concentration matrix $\Gamma$ has density:
$$
p_{\bar{X}, \Gamma}(X) = k \exp\left(-\frac{\Log_{\bar{X}}(X)^\top \Gamma\Log_{\bar{X}}(X)}{2}\right),
$$
where $k$ is a normalization constant without a closed form. The relation between $\Gamma$ and the covariance $\Sigma$ is given in Theorem 3 of \cite{pennecIntrinsicStatisticsRiemannian2006}. When $\Sigma = \sigma^2 I_d$, this reduces to the isotropic Gaussian of \cref{sec:isotropic_gaussian}. Another example of a Gaussian distribution of $\P_d$ is given in \cite{barbarescoGaussianDistributionsSpace2021}, where the author uses Souriau’s covariant Gibbs density to compute a Gaussian density of SPD matrices.

Other - non-Gaussian - probability distributions have been defined on the manifold $\P_d$ of SPD matrices. For example, a famous distribution used for SPD matrices is the Wishart distribution \cite{wishartGeneralisedProductMoment1928}. It is the distribution of sample covariance matrices of random vectors drawn from a multivariate Gaussian distribution and can be seen as a generalization of the gamma distribution to multiple dimensions. This Wishart distribution was first extended in \cite{ayadiTWDANovelDiscriminant2023} to the $t$-Wishart distribution similarly to the way the multivariate $t$-distribution extends the multivariate Gaussian. In \cite{ayadiEllipticalWishartDistributions2024}, the authors extend the Wishart distribution to an elliptical version, leading to a more robust and flexible distribution.
In \cite{ahandaCholeskyNormalDistribution2024}, the authors describe a Cholesky normal distribution on the manifold of SPD matrices. This distribution is related to the Wishart distribution as it relies on random vectors drawn from a multivariate Gaussian distribution. Moreover, the authors show that the Wishart distribution is approximately the Cholesky normal distribution for large degrees of freedom.


\section{Application to classification}
\label{sec:classification}
In this section, we show that popular classification algorithms on $\P_d$ can be reinterpreted in the light of a probabilistic framework using the different Gaussian distributions defined in \cref{sec:isotropic_gaussian}. In the following, we assume that we are in a supervised setting where we have $K$ classes each modeled by a distribution denoted $\alpha_k$. We start by introducing the Bayes Classifier (BC) :
\begin{definition}[Bayes Classifier (BC) \cite{bishop2007}]
    Given $K$ classes each modeled by a distribution $\alpha_k$, a Bayes Classifier (BC) assigns a new sample $Z$ to the class $k$ that maximizes the likelihood of $Z$ under the distribution $\alpha_k$.
\end{definition}

Our goal is to show that popular classification algorithms on $\P_d$ can be interpreted as BC using the different Gaussian distributions defined in \cref{sec:prob_distrib}. We will focus on the Minimum Distance to Mean (MDM) algorithm \cite{barachantRiemannianGeometryApplied2010}, on the Tangent Space Linear Discriminant Analysis (LDA) and on the Tangent Space Quadratic Discriminant Analysis (QDA) algorithms \cite{barachantMulticlassBrainComputer2012}. 

\subsection{Minimum Distance to Mean (MDM)}
The \emph{Minimum Distance to Mean} (MDM) algorithm is a simple and popular classification algorithm for SPD matrices described in \cite{barachantRiemannianGeometryApplied2010}. Given a training set of SPD matrices, this algorithm estimates the Riemannian mean $\hat{X}^k$ for each class $k \in \{1,...,K\}$ and assigns a new sample $Z$ to the class $k$ that minimizes the Riemannian distance to the estimated mean $\hat{X}^k$. We can reinterpret the MDM algorithm in the light of the isotropic Gaussian distribution defined in \cref{sec:isotropic_gaussian}. Indeed, we have the following:
\begin{proposition}
    Let us suppose that each class is modeled by an isotropic Gaussian distribution centered at $\bar{X}^k$ with a shared $\sigma$, i.e. $\alpha_k = G(\bar{X}^k, \sigma^2)$, then, the MDM converges to the BC when the number of data tends to infinity.
\end{proposition}
When the number of data tends to infinity, the estimation of the Riemannian mean of each class $\hat{X}^k$ converges to the true mean of the $k$-th class $\bar{X}^k$. Here, the spread $\sigma$ does not play any role in the classification, so one only needs to estimate the Riemannian mean $\bar{X}^k$ of each class.

\subsection{Tangent Space LDA/QDA}  
The Tangent Space LDA (resp. QDA) algorithm is a generalization of the LDA (resp. QDA) algorithm to the manifold of SPD matrices introduced in \cite{barachantMulticlassBrainComputer2012}. The idea is to first estimate the Riemannian mean $\hat{X}_N$ of the whole training set, and then project the training SPD matrices onto the tangent space $T_{\hat{X}_N} \P_d$. This tangent space being Euclidean, one finally applies the classical LDA (resp. QDA) algorithm (see section 4.3 of \cite{hastieElementsStatisticalLearning2009}) in this tangent space. Using the wrapped Gaussian distribution described in \cref{sec:wrapped_gaussian}, one has the following proposition:

\begin{proposition}
    When the number of data points tends to infinity, the Tangent Space LDA converges to a BC where the classes are modeled by wrapped Gaussian distributions centered at $$\bar{X} = \argmin_{Y \in \P_d} \int_{\P_d} \delta(X,Y)^2 \d \alpha(X)$$ where $\alpha = \frac{1}{k}\sum_{i=1}^k \alpha_k$ is the total distribution, and with a shared covariance matrix $\Sigma$, in other word, $\alpha_k = \WG(\bar{X}, \mu_k, \Sigma)$. Similarly, the Tangent Space QDA converges to a BC where the classes are modeled by $\alpha_k = \WG(\bar{X}, \mu_k, \Sigma_k)$.
\end{proposition}
Here, $\bar{X}$ is the Fr\'echet mean of total distribution as defined in definition 2 of \cite{pennecCurvatureEffectsEmpirical2019} and can be estimated by the Riemannian mean $\hat{X}_N$ of the training data. Then, estimating $\mu_k$ and $\Sigma_k$ boils down to the Euclidean estimation of the mean and covariance matrix in $T_{\hat{X}} \P_d$. One can use wrapped Gaussian to build new BC on $\P_d$ by modeling the classes $\alpha_k$ by $\WG(\bar{X}_k, \mu_k, \Sigma)$ or by $\WG(\bar{X}_k, \mu_k, \Sigma_k)$. This is done in \cite{desurrel2025wrappedgaussianmanifoldsymmetric}. These algorithms project each class to their own tangent space whereas only one tangent space is used in the Tangent Space LDA/QDA.
\subsection{Other probabilistic classifiers for SPD matrices}
In this paper, we mainly focus on the different Gaussian distributions described in \cref{sec:prob_distrib}. However, other authors have inquired BC based on other probability distributions. For instance, in \cite{ayadiEllipticalWishartDistributions2024}, the authors propose a BC based on the Elliptical Wishart distribution. They also adapt it to an unsupervised clustering scenario leveraging the K-means clustering algorithm. In \cite{ayadiTWDANovelDiscriminant2023}, the authors show that if the metric used in the MDM algorithm is based on the Kullback-Leibler divergence between two centered multivariate Gaussian distributions, then, the MDM algorithm is a BC based on the $t$-Wishart distribution.

\section{Application to outlier detection: Riemannian potato}
\label{sec:outlier_detection}
Another tool that is used when the SPD matrices are covariance matrices of ElectroEncephaloGraphy (EEG) signals is the \emph{Riemannian potato} introduced in \cite{barachantRiemannianPotatoAutomatic2013}. This tool is used to automatically detect outliers in the data. The idea of the Riemannian potato is to estimate a reference SPD matrix and a measure of dispersion (z-score), and then to reject all SPD matrices that are too far from the reference matrix. Let $(X_i)_{i = 1,...,N}$ be a set of SPD matrices, and let us denote $\bar{X}$ the reference SPD matrix. Then, the z-score $z$ of $X \in \P_d$ is computed as:
$$z = \frac{\delta(X, \bar{X}) - \mu}{\sigma} \text{ where } \mu = \frac{1}{N}\sum_{i=1}^N \delta(X_i, \bar{X}) \text{ and } \sigma = \sqrt{\frac{1}{N}\sum_{i=1}^N (\delta(X_i, \bar{X})- \mu)^2}.$$
Provided a threshold $z_{\text{th}}$, the matrix $X$ is accepted if $z \leq z_{\text{th}}$. The Riemannian potato is a simple yet efficient tool to detect outliers in the data. In practice, the reference matrix $\bar{X}$ is a Riemannian mean computed iteratively as new samples are added to the dataset. Using the isotropic Gaussian distribution defined in \cref{sec:isotropic_gaussian}, we can see that the Riemannian potato rejects points that are too unlikely under a certain isotropic Gaussian:
\begin{proposition}
    Let $\bar{X} \in \P_d$ be a reference matrix, $\mu \in \R, \sigma > 0$ and $z_{\text{th}} > 0$. Then, $X \in \P_d$ is accepted by the Riemannian potato with threshold $z_{\text{th}}$ if and only if $\mathcal{L}_{\bar{X}, \sigma}(X) \geq \ell_{\text{th}}$ where $\mathcal{L}_{\bar{X}, \sigma}$ denotes the likelihood of $G(\bar{X}, \sigma)$ and where
    $$\ell_{\text{th}} = \exp\left(-\frac{1}{2}\left(z_{\text{th}} + \frac{\mu}{\sigma}\right)\right)$$
\end{proposition}

Riemannian potato have been extended to a Riemannian potato field \cite{barthelemyRiemannianPotatoField2019}. Another use of the isotropic Gaussian distribution to detect outliers is done in \cite{wangNonparametricOnlineChange2024} where the author built an online change detection algorithm based on the estimation of the Riemannian mean of the data before and after the change.

\section{Application to dimension reduction}
\label{sec:dim_reduction}

\begin{figure}[t]
    \centering
    \includegraphics[width=\linewidth]{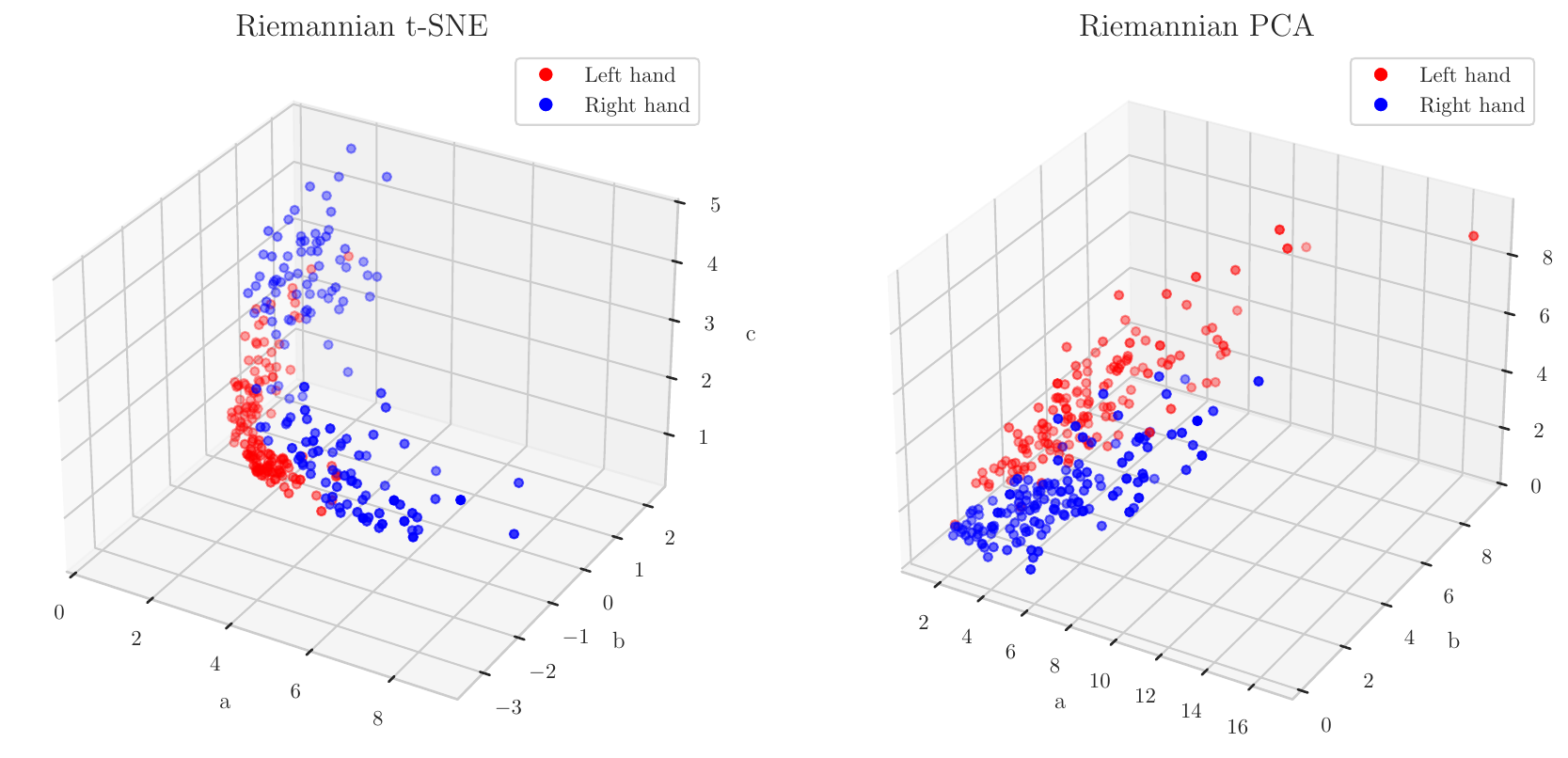}
    \caption{Results of the Riemannian t-SNE and PCA algorithms applied to the covariance matrices of EEG signals from subject 8 of the BNCI2014001 \cite{BNCI2014004} dataset. The points are colored according to the class of the task performed by the subject. Both algorithms reduce the data to $2 \times 2$ SPD matrices, which can be visualized in 3D. Indeed, a $2 \times 2$ SPD matrices is of the form $\begin{pmatrix} a & b \\ b & c \end{pmatrix}$, which can be represented in 3D by the point $(a, b, c)$. }
    \label{fig:riemannian_dimension_reduction}
\end{figure}

\subsection{Riemannian t-SNE}
In \cite{surrel2025geometryaware}, the authors define a Riemannian version of the t-SNE algorithm from \cite{maatenVisualizingDataUsing2008} that reduces $d \times d$ SPD matrices into $2 \times 2$ SPD matrices. As the set $\P_2$ of $2 \times 2$ SPD matrices is of dimension $3$, it can be visualized in a 3D space. To adapt the algorithm to the manifold $\P_d$, the authors replaced the Euclidean distance by the AIRM distance and showed that the algorithm is still valid. The Riemannian t-SNE algorithm is based on the computation of similarities between the high-dimensional SPD matrices using a Gaussian kernel. This kernel is based on the isotropic Gaussian distribution defined in \cref{sec:isotropic_gaussian}. Denoting $X_1,...,X_N$ the set of high-dimensional SPD matrices, the similarity $p_{i\mid j}$ of the matrix $X_i$ to $X_j$ is the conditional probability $p_{j|i}$ that $X_i$ would pick $X_j$ as its neighbor if they were picked in proportion to their probability density under an isotropic Gaussian $G(X_i, \sigma_i^2)$ centered at $X_i$. The dispersion $\sigma_i$ is computed using the perplexity parameter of the t-SNE algorithm. Then, the Riemannian t-SNE aims at learning $Y_1,...,Y_N \in \P_2$, i.e., the SPD matrices with the reduced dimension, that reflect the similarities $p_{i \mid j}$ as well as possible. For this, the joint probabilities $q_{ij}$  of the low-dimensional SPD matrices $Y_i$ and $Y_j$ are computed using a Riemannian version of the Student-t distribution with one degree of freedom. Finally, the Kullback-Leibler divergence between the two distributions is minimized using a Riemannian gradient descent algorithm. 

We apply this Riemannian t-SNE to a dataset of SPD matrices that are covariance matrices of EEG signals used in Brain Computer Interfaces. The dataset is the BNCI2014001 dataset \cite{BNCI2014004} from MOABB \cite{Aristimunha_Mother_of_all_2023} that contains 9 subjects performing motor imagery tasks. The Riemannian t-SNE algorithm is applied to the covariance matrices of the EEG signals of subject 8 of the dataset, and the results of given at \cref{fig:riemannian_dimension_reduction}. The Riemannian t-SNE algorithm reduces the data to $2 \times 2$ SPD matrices, which can be visualized in 3D. The points are colored according to the class of the task performed by the subject. The Riemannian t-SNE algorithm is able to separate the different classes of tasks performed by the subject, showing that it is a good tool for dimension reduction on $\P_d$.

\subsection{Riemannian PCA}
The goal of the \emph{Principal Component Analysis} (PCA) algorithm is to project the data onto a lower-dimensional space while maximizing the variance of the projected points. In \cite{horevGeometryawarePrincipalComponent2017}, the authors extend the PCA to a Riemannian setting of SPD matrices. Given a set $X_1,...,X_N$ of SPD matrices, their goal is to find a $W$ that maximizes the variance of the projected points $W^\top X_1W,...,W^\top X_NW$. The matrix $W$ is chosen in $\mathcal{G}(d,p)$, the Grassmann manifold, that is the manifold of $d \times p$ matrices of rank $p$. This is to make sure that the projected points are SPD matrices. Therefore, the Riemannian PCA is defined as the following optimization problem, where $\hat{X}_N$ is the Riemannian mean of the set $X_1,...,X_N$:
$$W \in \underset{W \in \mathcal{G}(d,p)}{\text{argmax}} \sum_{i=1}^N \delta(W^\top X_i W, W^\top \hat{X}_N W)^2$$

We recall that the variance of a distribution $\mu$  on $\P_d$, at a point $X$, as defined in \cite{pennecCurvatureEffectsEmpirical2019}, is $\sigma^2(X) = \int_{\P_d} \delta(X,Y)^2 \d \mu(Y)$. In the discrete setting of Riemannian PCA, the integral is replaced by a sum. Therefore, the underlying distribution of the Riemannian PCA is the isotropic Gaussian distribution defined in \cref{sec:isotropic_gaussian}. 

We also apply this Riemannian PCA algorithm to the same dataset of covariance matrices of EEG signals used in the Riemannian t-SNE algorithm. The results are given at \cref{fig:riemannian_dimension_reduction}. As the Riemannian t-SNE, we reduce the covariance matrices to $2 \times 2$ SPD matrices, which can be visualized in 3D. The Riemannian PCA algorithm also demonstrates the ability to separate the different classes of tasks performed by the subject, confirming its effectiveness for dimension reduction on $\P_d$.

\section{Conclusion}
In this paper, we first demonstrated how the isotropic Gaussian distribution and the wrapped Gaussian distribution can be used to reinterpret the MDM, Tangent Space LDA and Tangent Space QDA classifiers as Bayes Classifiers. Therefore, these algorithms share a common probabilistic framework, differing only in their choice of data distribution. Using the same framework, we showed that the Riemannian potato rejects points that are unlikely under an isotropic Gaussian distribution. Finally, the Riemannian t-SNE and the Riemannian PCA algorithms were also encompassed in the same probabilistic framework. This work shows that the various tools used on the manifold of SPD matrices can be brought together under the same probabilistic framework.

This work opens the door to using other probability distributions on $\P_d$ and constructing other ML tools using them. For example, a possible use of the different Gaussian distributions is the construction of Gaussian kernels on $\P_d$. Kernels on $\P_d$ have already been investigated \cite{YgerReviewKernelsCovariance2013,jayasumana2014kernelmethodsriemannianmanifold}, and the use of the isotropic Gaussian distribution to build a kernel leads to a strong limitation due to the curvature of $\P_d$ (see \cite{feragenGeodesicExponentialKernels2015}). However, the wrapped Gaussian distribution could lead to a more flexible kernel that could be used in a wide range of applications. Deep Learning methods have also been investigated on $\P_d$ \cite{huang2017riemannian,liSPDDDPMDenoisingDiffusion2023,wangDreamNetDeepRiemannian2023}, and the use of the different probability distributions could lead to a new outlook on deep models on $\P_d$. On more general Riemannian manifolds, the isotropic Gaussian has for example been used to extend Variational Flow Matching to Riemannian manifolds in \cite{zaghen2025variationalflowmatchinggeneral} and wrapped distributions have been leveraged to learn Riemannian latent spaces in \cite{rozo2025riemann}.

\subsubsection{\ackname}
\begin{credits}
This work was funded by the French National Research Agency for project PROTEUS (grant ANR-22-CE33-0015-01). Sylvain Chevallier is supported by DATAIA (ANR-17-CONV-0003). 
\end{credits}
%


%
%
\bibliographystyle{splncs04}
\bibliography{biblio}

\end{document}